\documentclass[10pt,twocolumn,letterpaper]{article}
\pdfoutput=1
\usepackage{cvpr}
\usepackage{times}
\usepackage{epsfig}
\usepackage{graphicx}
\usepackage{amsmath}
\usepackage{amssymb}
\usepackage{multirow}
\usepackage{paralist}
\usepackage{makecell}
\usepackage[english]{babel}

\usepackage{authblk}

\usepackage[pagebackref=true,breaklinks=true,letterpaper=true,colorlinks,bookmarks=false]{hyperref}

\cvprfinalcopy 

\def\cvprPaperID{2256} 

\ifcvprfinal\fi
\begin{document}

\title{MLCVNet: Multi-Level Context VoteNet for 3D Object Detection}

\author[1]{Qian Xie}
\author[2]{Yu-Kun Lai}
\author[2]{Jing Wu}
\author[1]{Zhoutao Wang}
\author[1]{Yiming Zhang}
\author[3]{Kai Xu}
\author[1]{Jun Wang \thanks{Corresponding author: wjun@nuaa.edu.cn}}
\affil[1]{Nanjing University of Aeronautics and Astronautics}
\affil[2]{Cardiff University}
\affil[3]{National University of Defense Technology}

\maketitle
\thispagestyle{empty}

\begin{abstract}
  In this paper, we address the 3D object detection task by capturing multi-level contextual information with the self-attention mechanism and multi-scale feature fusion.
  Most existing 3D object detection methods recognize objects individually, without giving any consideration on contextual information between these objects.
  Comparatively, we propose Multi-Level Context VoteNet (MLCVNet) to recognize 3D objects correlatively, building on the state-of-the-art VoteNet.
  We introduce three context modules into the voting and classifying stages of VoteNet to encode contextual information at different levels.
  Specifically, a Patch-to-Patch Context (PPC) module is employed to capture contextual information between the point patches, before voting for their corresponding object centroid points.
  Subsequently, an Object-to-Object Context (OOC) module is incorporated before the proposal and classification stage, to capture the contextual information between object candidates.
  Finally, a Global Scene Context (GSC) module is designed to learn the global scene context.
  We demonstrate these by capturing contextual information at patch, object and scene levels.
  Our method is an effective way to promote detection accuracy, achieving new state-of-the-art detection performance on challenging 3D object detection datasets, i.e., SUN RGBD and ScanNet.
  We also release our code at https://github.com/NUAAXQ/MLCVNet.
\end{abstract}

\section{Introduction}
3D object detection is becoming an active research topic in both computer vision and computer graphics.
Compared to 2D object detection in RGB images, predicting 3D bounding boxes in real world environments captured by point clouds is more essential for many tasks~\cite{song2016deep} such as indoor robot navigation~\cite{mccormac2018fusion++}, robot grasping~\cite{wang2019densefusion}, etc.
However, the unstructured data in point clouds makes the detection more challenging than in 2D. In particular, the popular convolutional neural networks (CNNs), which are highly successful in 2D object detection, are difficult to be applied to point clouds directly.

\begin{figure}[!t]
  \centering
  \includegraphics[width=0.95\linewidth]{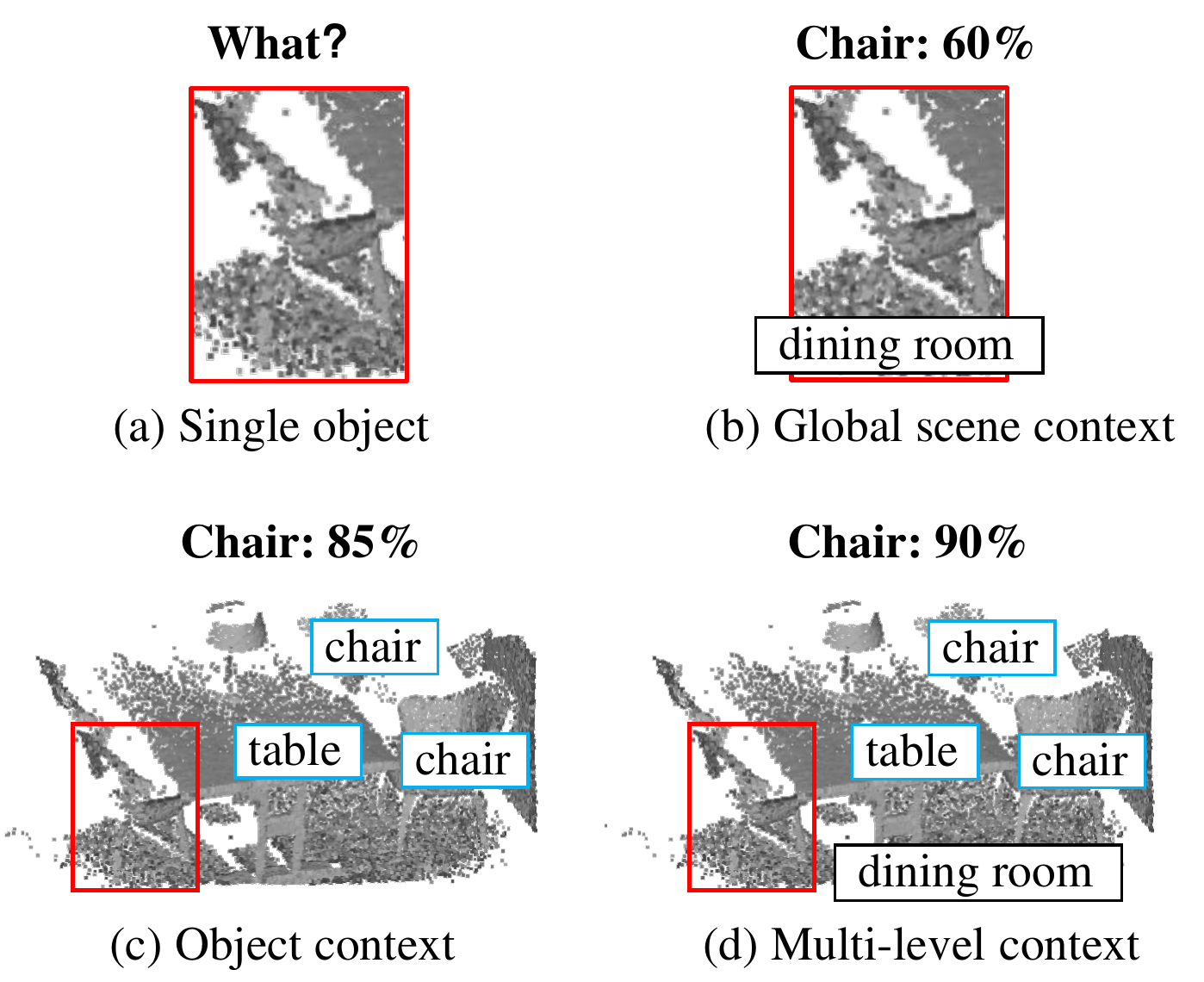}
  \caption{Illustration of the importance of multi-level contextual information for 3D object detection from point cloud data. (a) It is hard to recognize the object when the point cloud is shown independently. (b)-(d) When the surrounding environment information is given, we can then recognize the chair easily.
  \textit{In fact, unlike general object detection in open scenes, indoor scenes usually contain strong context constraints, which can be utilized in indoor scene understanding tasks such as 3D object detection.}
  \label{fig:context_illu}}
\end{figure}

Growing interests have been attracted to tackle this challenge.
With the emergence of deep 3D points processing networks, such as~\cite{qi2017pointnet,qi2017pointnet++}, several deep learning based 3D object detection works have been proposed recently to detect objects directly from 3D point clouds~\cite{hou20193d,qi2019deep}.
The most recent work VoteNet~\cite{qi2019deep} proposed an end-to-end 3D object detection network on the basis of Hough voting.
VoteNet transfers the traditional Hough voting procedure into a regression problem implemented by a deep network, and samples
a number of seed points from the input point cloud to generate patches voting for potential object centers.
The voted centers are then used to estimate the 3D bounding boxes.
The voting strategy enables VoteNet to significantly reduce the searching space and achieve the state-of-the-art results in several benchmark datasets.
However, treating every point patch and object individually, VoteNet lacks the consideration of the relationships between different objects and between objects and the scene they belong to, which limits its detection accuracy.

\begin{figure}[!t]
  \centering
  \includegraphics[width=\linewidth]{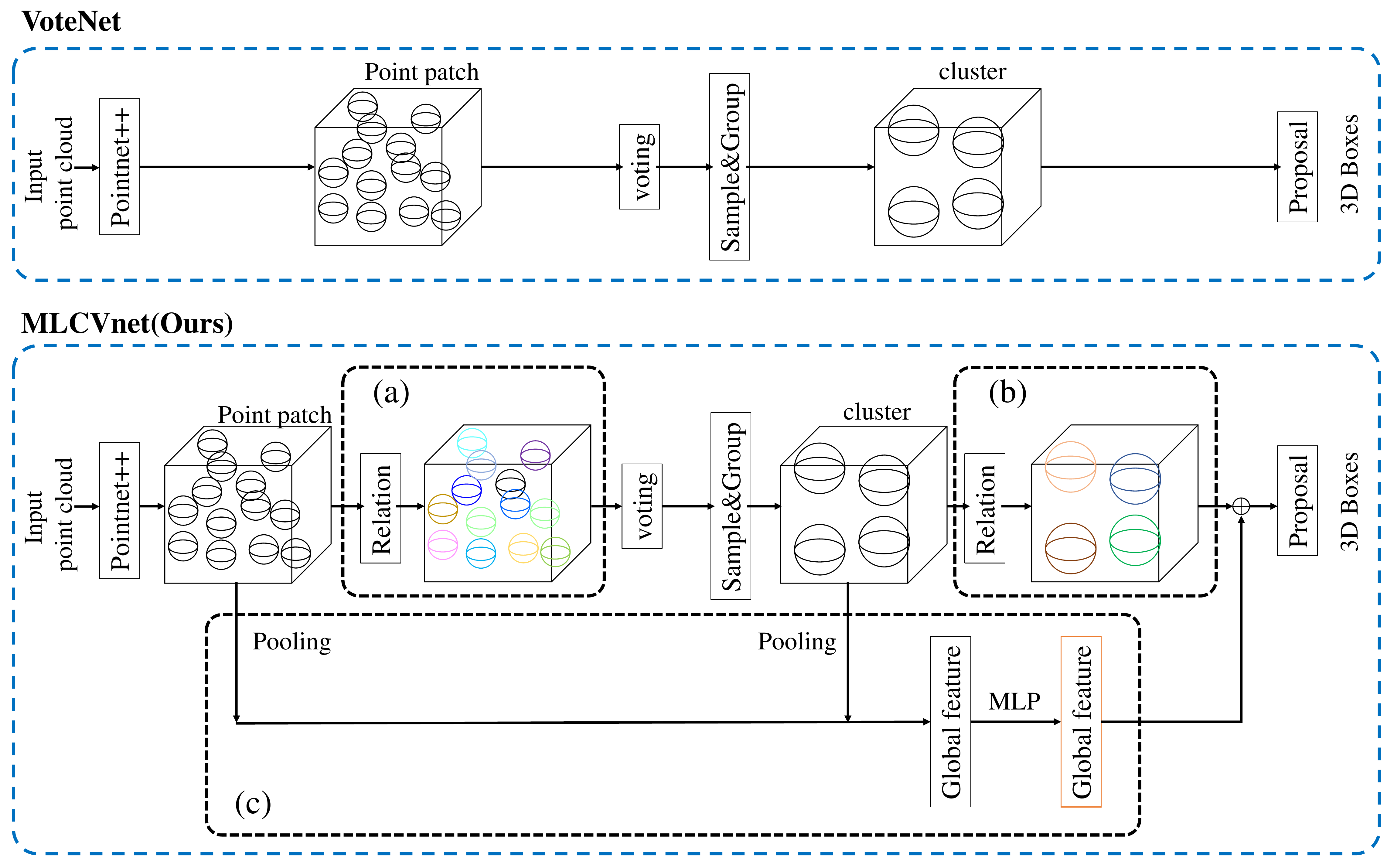}
  \caption{Comparison of architectures between VoteNet~\cite{qi2019deep} and the proposed MLCVNet.
  Three sub-modules are integrated to capture the multi-level contextual information in point cloud data.
  (a) patch level context module; (b) object level context module; (c) global scene context module.
  \label{fig:votenet_vs_mlcvnet}}
\end{figure}

An example can be seen in Fig.~\ref{fig:context_illu}. Point clouds, captured by e.g. depth cameras,
often contain noisy and missing data. This together with indoor occlusions makes it difficult even for humans to recognize what and where an object is in Fig.~\ref{fig:context_illu}(a). Nevertheless, considering the surrounding contextual information in Figs.~\ref{fig:context_illu}(b-d), it is much easier to recognize it is a chair given the surrounding chairs and the table in the dining room scene.
Actually, the representation of a scanned point set could be ambiguous when it is presented individually,
due to lack of color appearance and data missing problems.
Therefore, we argue that indoor depth scans are often so occluded that contexts could even play a more important role in recognizing objects than the point data itself.
This contextual information has been demonstrated to be helpful in a variety of computer vision tasks, including object detection~\cite{hu2018relation,yu2016role}, image semantic segmentation~\cite{zhang2019co,fu2019dual} and 3D scene understanding~\cite{zhang2014panocontext,zhang2017deepcontext}.
In this paper, we show how to leverage the contextual information in 3D scenes to boost the performance of 3D object detection from point clouds.

In our view, contextual information for 3D object detection consists of multiple levels.
At the lowest is the patch level where the data missing problem is mitigated with a weighted sum over similar point patches to assist more accurate voting of object centers.
At the object level, coexistence of objects provides strong hints on detection of certain objects.
For example, as shown in Fig.~\ref{fig:context_illu}(d), the detected table can give a tendency for chairs to be detected at surrounding points.
At the scene level, global scene clues can also prevent an object from being detected in an improper scene.
For example, we will not expect to detect a bed in a kitchen.
The contexts at different levels complement each other and are utilized together to assist the correct inference of objects in noisy and cluttered environments.

\begin{figure*}[!t]
  \centering
  \includegraphics[width=\linewidth]{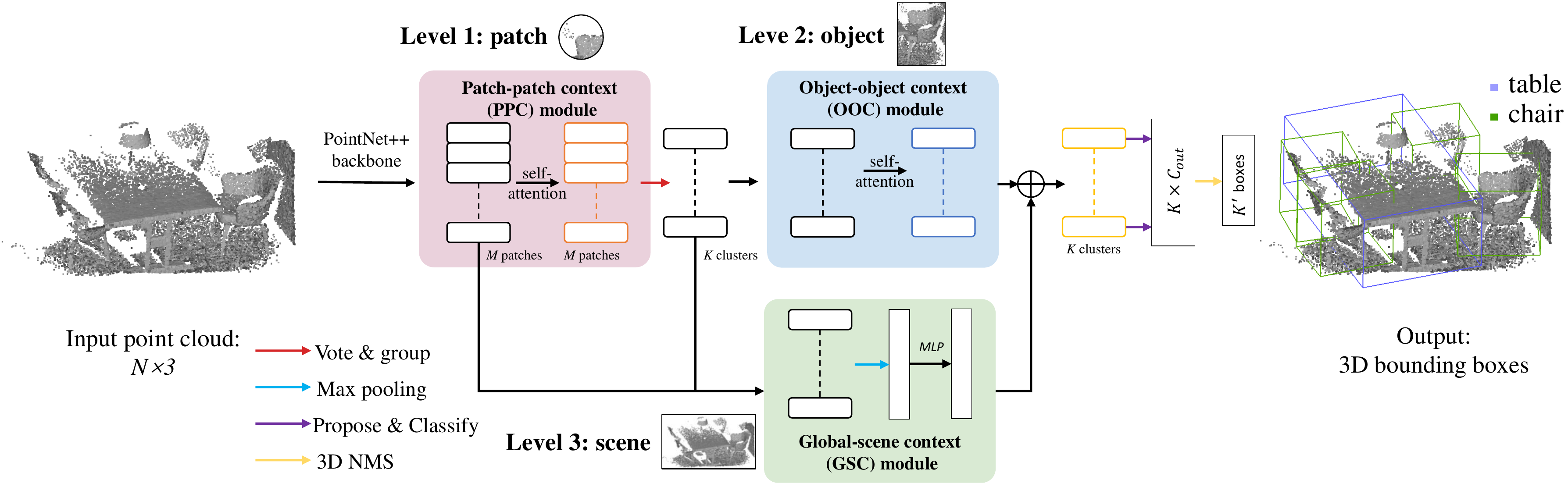}
  \caption{Architecture of the proposed MLCVNet for 3D object detection in point cloud data. Three new sub-modules are proposed to capture the multi-level contextual information in 3D indoor scene object detection.
  \label{fig:network}}
\end{figure*}

We thus propose a novel 3D object detection framework, called Multi-Level Context VoteNet (MLCVNet), to incorporate into VoteNet
multi-level contextual information for 3D object detection.
Specifically, we propose a unified network to model the multi-level contexts, from local point patches to global scenes.
The difference between VoteNet and the proposed network is highlighted in Fig.~\ref{fig:votenet_vs_mlcvnet}.
To model the contextual information, three sub-modules are proposed in the framework, i.e., patch-to-patch context (PPC) module, object-to-object context (OOC) module and the global scene context (GSC) module.
In particular, similar to~\cite{zhang2019pcan}, we use the self-attention mechanism to model the relationships between elements in both PPC and OOC modules.
These two sub-modules aim at adaptively encoding contextual information at the patch and object levels, respectively.
For the scene-level, we design a new branch as shown in Fig.~\ref{fig:votenet_vs_mlcvnet}(c) to fuse multi-scale features to equip the network with the ability of learning global scene context.
In summary, the contributions of this paper include:
\begin{compactitem}
\item We propose the first 3D object detection network that exploits \emph{multi-level} contextual information at patch, object and global scene levels.
\item We design three sub-modules, including two self-attention modules and a multi-scale feature fusion module, to capture the contextual information at multiple levels in 3D object detection. The new modules nicely fit in the state-of-the-art VoteNet framework. Ablation study demonstrates the effectiveness of these modules in improving detection accuracy.
\item Extensive experiments demonstrate the benefits of multi-level contextual information. The proposed network outperforms state-of-the-art methods on both SUN RGB-D and ScanNetV2 datasets.
\end{compactitem}

\section{Related Work}
\label{related_work}
\subsection{3D Object Detection From Point Clouds}
Object detection from 2D images has been studied for decades.
Since the development of deep convolutional neural networks (DCNNs)~\cite{krizhevsky2012imagenet}, both the accuracy and efficiency of 2D object detection have been significantly improved by deep learning techniques~\cite{girshick2015fast,ren2015faster}.
Compared to 2D, 3D object detection was dominated by non-deep learning based methods~\cite{nan2012search,li2015database,wang2016cluttered} until the recent couple of years.
With the development of deep learning on 3D point clouds~\cite{wang2017cnn,li2018pointcnn,atzmon2018point}, many deep learning based 3D object detection architectures have emerged~\cite{chen2016monocular,chen2017multi,lahoud20172d}.
However, most of these methods depend on using 2D detectors as an intermediate step, which restricts their generalization to situations where 2D detectors do not work well~\cite{qi2018frustum}.
To address this issue, several deep learning based 3D detectors  which directly take raw point clouds as input have been proposed  recently~\cite{zhou2018voxelnet,yang2019learning,hou20193d}.
In~\cite{shi2019pointrcnn}, the authors introduced a two-stage 3D object detector, PointRCNN.
Their method first generates several 3D bounding box proposals, and then refines these proposals to obtain the final detection results.
Instead of directly treating 3D object proposal generation as a bounding box regression problem, in~\cite{yi2019gspn}, a novel 3D object proposal approach was proposed by taking an analysis-by-synthesis strategy and reconstructing 3D shapes from point clouds.
Inspired by the Hough voting strategy for 2D object detection in~\cite{leibe2004combined}, the work in~\cite{qi2019deep} presents an end-to-end trainable 3D object detection network, which directly deals with 3D point clouds, by virtue of the huge success in PointNet/PointNet++~\cite{qi2017pointnet,qi2017pointnet++}.
Although a lot of methods have been proposed recently, there is still large room for improvement especially for real-world challenging cases.
Previous works largely ignored contextual information, i.e., relationships within and between objects and scenes.
In this work, we show how to leverage the contextual information to improve the accuracy of 3D object detection.

\subsection{Contextual Information}
The work in~\cite{mottaghi2014role} has demonstrated that contextual information has significant positive effect on 2D semantic segmentation and object detection.
Since then, contextual information has been successfully employed to improve performance on many tasks such as 2D object detection~\cite{yu2016role,hu2018relation,liu2018structure}, 3D point matching~\cite{deng2018ppfnet}, point cloud semantic segmentation~\cite{engelmann2017exploring,ye20183d}, and 3D scene understanding~\cite{zhang2014panocontext,zhang2017deepcontext}.
The work in~\cite{hu2018semantic} achieves reasonable results on instance segmentation of 3D point clouds via analyzing point patch context.
In~\cite{shi2019hierarchy}, a recursive auto-encoder based approach is proposed to predict 3D object detection via exploring hierarchical context priors in 3D object layout.
Inspired by the self-attention idea in natural language processing~\cite{vaswani2017attention}, recent works connect the self-attention mechanism with contextual information mining to improve scene understanding tasks such as image recognition~\cite{hu2018squeeze}, semantic segmentation~\cite{fu2019dual} and point cloud recognition~\cite{xie2018attentional}.
As to 3D point data processing, the work in~\cite{zhang2019pcan} proposes to utilize the attention network to capture the contextual information in 3D points.
Specifically, it presents a point contextual attention network to encode local features into a global descriptor for point cloud based retrieval.
In~\cite{paigwar2019attentional}, an attentional PointNet is proposed to search regions of interest instead of processing the whole input point cloud, when detecting 3D objects in large-scale point clouds.
Different from previous works, we are interested in exploiting the combination of \emph{multi-level} contextual information for 3D object detection from point clouds.
In particular, we integrate two self-attention modules and one multi-scale feature fusion module into a deep Hough voting network to learn multi-level contextual relationships between patches, objects and the global scene.

\section{Approach}
\label{method}
As shown in Fig.~\ref{fig:network}, our MLCVNet contains four main components: a fundamental 3D object detection framework based on VoteNet
which follows the architecture in~\cite{qi2019deep}, and three context encoding modules.
The PPC (patch-patch context) module
combines the point groups to encode the patch correlation information, which helps to vote for more accurate object centers.
The OOC (object-object context) module
is for capturing the contextual information between object candidates.
This module helps to improve the results of 3D bounding box regression and classification.
The GSC (global scene context) module
is to integrate the global scene contextual information.
In brief, the proposed three sub-modules are designed to capture complementary contextual information in 3D object detection at multiple levels, with the aim to improve the detection performance in 3D point clouds.

\subsection{VoteNet}
\label{Vote}
VoteNet~\cite{qi2019deep} is the baseline of our work.
As illustrated in Fig.~\ref{fig:votenet_vs_mlcvnet}, it is an end-to-end trainable 3D object detection network consisting of three main blocks: \emph{point feature extraction}, \emph{voting}, and \emph{object proposal and classification}.

\begin{figure}[!t]
  \centering
  \includegraphics[width=\linewidth]{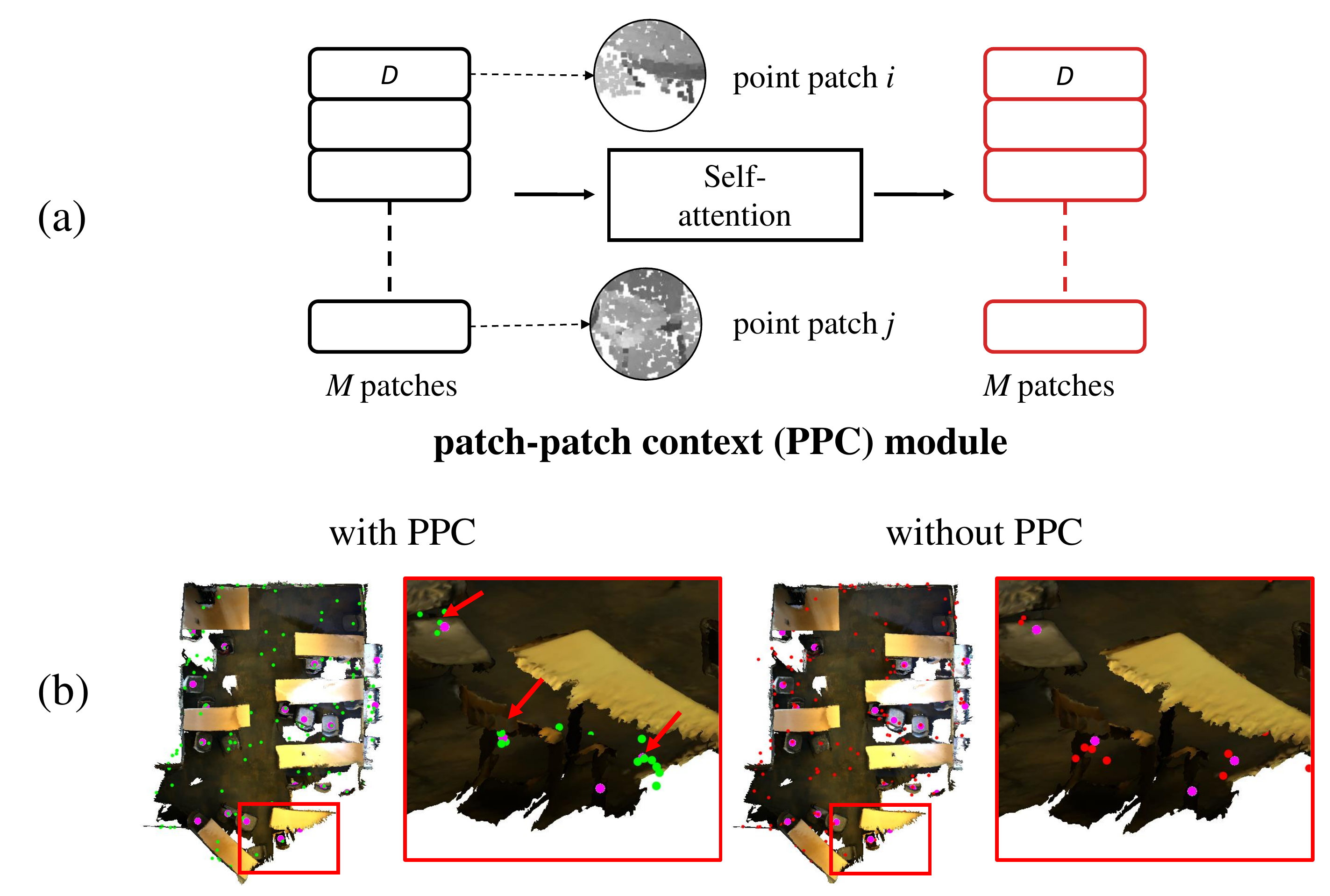}
  \caption{(a) Architecture details of the PPC module.
  CGNL~\cite{yue2018compact} is adopted as the self-attention module in our paper.
  (b) Comparison of center voting results with (green) and without (red) the PPC module. Pink points denote annotated ground-truth centers.
  \label{fig:ppc}}
\end{figure}

To extract point features, PointNet++ is used as the backbone network for seed sampling and extracting high dimensional features for the seed points from the raw input point cloud.
The features of each seed point contain information from its surrounding points within a radius as illustrated in Fig.~\ref{fig:ppc}(a).
Analogous to regional patches in 2D, we thus call these seed points \textit{point patches} in the remaining of this paper.
The voting block takes the point patches with extracted features as input and regresses object centers.
This center point prediction is performed by a multi-layer perceptron  (MLP) which simulates the Hough voting procedure.
Clusters are then generated by grouping the predicted centers, and form object candidates, from which the 3D bounding boxes are then proposed and classified through another MLP layer.

Note that in VoteNet, both the point patches and the object candidates are processed independently, ignoring the surrounding patches or objects.
However, we argue that relationships between these elements (i.e., point patches and object candidates) are useful information for object detection.
Thus, we introduce our MLCVNet
to encode these relationships.
Our detection network follows the general framework
of VoteNet,
but integrates three new sub-modules to capture multi-level contextual information.

\subsection{PPC Module}
\label{PPC}
We consider relationships between point patches as the first level of context, i.e., patch-patch context (PPC), as shown in Fig.~\ref{fig:ppc}(a).
At this level, contextual information between point patches,
on the one hand, helps relieve the data missing problem via gathering supplementary information from similar patches.
On the other hand, it considers inter-relationships between patches for voting~\cite{wang2013learning} by aggregating voting information from both the current point patch and all the other patches.
We thus propose a sub-network, PPC module, to capture the relationships between point patches.
For each point patch, the basic idea is to employ a self-attention module to aggregate information from all the other patches before sending it to the voting stage.

As shown in Fig.~\ref{fig:ppc}(a), after feature extraction using PointNet++, we get a feature map $\mathbf{A} \in \mathbb{R}^{1024 \times D}$, where $1024$ is the number of point patches sampled from the raw point cloud, and $D$ is the dimension of the feature vector.
We intend to generate a new feature map $\mathbf{A'}$ that encodes the correlation between any two point patches,
and it can be formulated as the non-local operation:
\begin{equation}\label{NL_formula}
    \mathbf{A'}=f(\theta(\mathbf{A}), \phi(\mathbf{A})) g(\mathbf{A})
\end{equation}
where $\theta(\cdot), \phi(\cdot), g(\cdot)$ are three different transform functions, and $f(\cdot, \cdot)$ encodes the similarities between any two positions of the input feature.
Moreover, as shown in~\cite{hu2018squeeze}, channel correlations in the feature map also contribute to the contextual information modeling in object detection tasks,
we thus make use of the compact generalized non-local network (CGNL)~\cite{yue2018compact} as the attention module to explicitly model rich correlations between any pair of point patches and of any channels in the feature space.
CGNL requires light computation and little additional parameters, making it more practically applicable.
After the attention module, each row in the new feature map still corresponds to a point patch, but contains not only its own local features, but also the information associated with all the other point patches.

The effectiveness of the PPC module is visualized in Fig.~\ref{fig:ppc}(b).
As shown, with the PPC module, the voted centers are more meaningful with more of them appearing on objects rather than on non-object regions.
Moreover, the voted centers are more closely clustered compared to those without the module.
The results demonstrate that our self-attention based weighted fusion over local point patches can enhance the performance of voting for object centers.

\begin{figure}[!t]
  \centering
  \includegraphics[width=\linewidth]{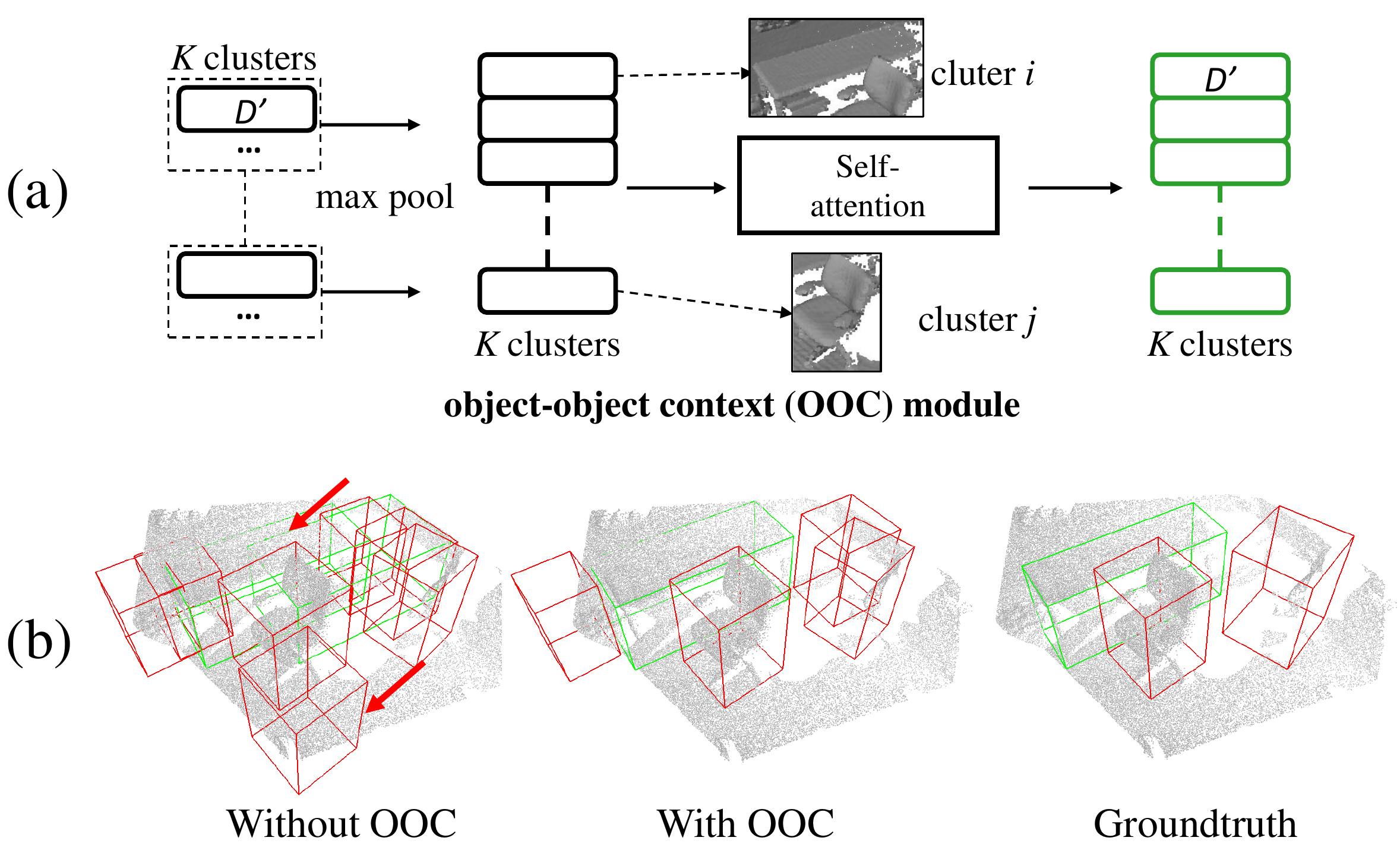}
  \caption{(a) Architecture details of the OOC module. CGNL~\cite{yue2018compact} is adopted as the self-attention module.
  (b) Comparison of results with and without the OOC module.
  \label{fig:ooc}}
\end{figure}

\subsection{OOC Module}
\label{OOC}
Most existing object detection frameworks detect each object individually.
VoteNet is no exception, where each cluster is independently fed into the MLP layer to regress its object class and bounding box.
However, combining features from other objects gives more information on the object relationships, which has been demonstrated to be helpful in image object detection~\cite{chen2018context}.
Intuitively, objects will get weighted messages from those highly correlated objects.
In such a way, the final predicted object result is not only determined by its own individual feature vector but also affected by object relationships.
We thus regard the relationships between objects as the second level contextual information, i.e., object-object context (OOC).

We get a set of vote clusters $\mathbf{C}=\left\{\mathcal{C}_{1}, \mathcal{C}_{2}, \dots, \mathcal{C}_{K}\right\}$ after grouping the voted centers.
$K$ is the number of generated clusters in this work.
Each cluster $\mathcal{C}=\left\{v_{1}, v_{2}, \dots, v_{n}\right\}$ is fed into an MLP followed by a max pooling to form a single vector representing the cluster. Here $v_i$ represents the $i$-th vote in $\mathcal{C}$, and $n$ is the number of votes in $\mathcal{C}$.
Then comes the difference from VoteNet. Instead of processing each cluster vector independently to generate a proposal and classification, we consider the relationships between objects.
Specifically, we introduce a self-attention module before the proposal and classification step, as shown in Fig.~\ref{fig:network} (the blue module). Fig.~\ref{fig:ooc}(a) shows the details inside the OOC module.
Specifically, after max pooling, the cluster vectors $\textbf{{C}}\in\mathbb{R}^{K\times D'}$
are fed into the CGNL attention module to generate a new feature map to record the affinity between all clusters.
The encoding of object relationships can be summarized as:
\begin{equation}\label{OOC_formula}
  \mathcal{C}_{OOC}=Attention(\max _{i=1, \ldots, n}\left\{MLP\left(v_i\right)\right\})
\end{equation}
where $\mathcal{C}_{OOC}$ is the enhanced feature vector in the new feature map $\textbf{{C}}_{OOC}\in \mathbb{R}^{K\times D'}$, and $Attention(\cdot)$ is the CGNL attention mapping.
By doing so, the contextual relationships between these clusters (objects) are encoded into the new feature map.

The effectiveness of the OOC module is visualized in  Fig.~\ref{fig:ooc}(b).
As shown, with the OOC module, there are fewer detected objects overlapping with each other, and the positions of the detected objects are more accurate.

\begin{table*}
\centering
\resizebox{\textwidth}{15mm}{
\begin{tabular}{c|c|*{10}{p{0.8cm}<{\centering}}|c}
\hline
           & Input    & table         & sofa          & booksh     & chair         & desk          & dresser       & nightst    & bed           & bathtub       & toilet        & mAP@0.25      \\ \hline
DSS~\cite{song2016deep}           & Geo+RGB  & 50.3          & 53.5          & 11.9          & 61.2          & 20.5          & 6.4           & 15.4          & 78.8          & 44.2          & 78.9          & 42.1          \\
2D-driven~\cite{lahoud20172d}     & Geo+RGB  & 37.0            & 50.4          & 31.4          & 48.3          & 27.9          & 25.9          & 41.9          & 64.5          & 43.5          & 80.4          & 45.1          \\
COG~\cite{ren2016three}           & Geo+RGB  & \textbf{51.3}          & 51.0          & 31.8          & 62.2          & \textbf{45.2}          & 15.5          & 27.4          & 63.7          & 58.3          & 70.1          & 47.6          \\
F-PointNet~\cite{qi2018frustum}    & Geo+RGB  & 51.1          & 61.1          & \textbf{33.3 }         & 64.2          & 24.7          & \textbf{32.0}          & 58.1          & 81.1          & 43.3          & 90.9          & 54.0          \\
VoteNet~\cite{qi2019deep}       & Geo-only  & 47.3          & 64.0          & 28.8          & 75.3          & 22.0          & 29.8          & \textbf{62.2}          & 83.0          & 74.4          & \textbf{90.1}          & 57.7          \\\hline
MLCVNet(ours) & Geo only & 50.4 & \textbf{66.3} & 31.9 & \textbf{75.8} & 26.5 & 31.3 & 61.5 & \textbf{85.8} & \textbf{79.2} & 89.1 & \textbf{59.8} \\ \hline
\end{tabular}
}
\smallskip
\caption{Performance comparison with state-of-the-art 3D object detection networks on SUN RGB-D V1 validation set.
\label{tab:sunrgbd}}
\end{table*}

\begin{figure}[!t]
  \centering
  \includegraphics[width=\linewidth]{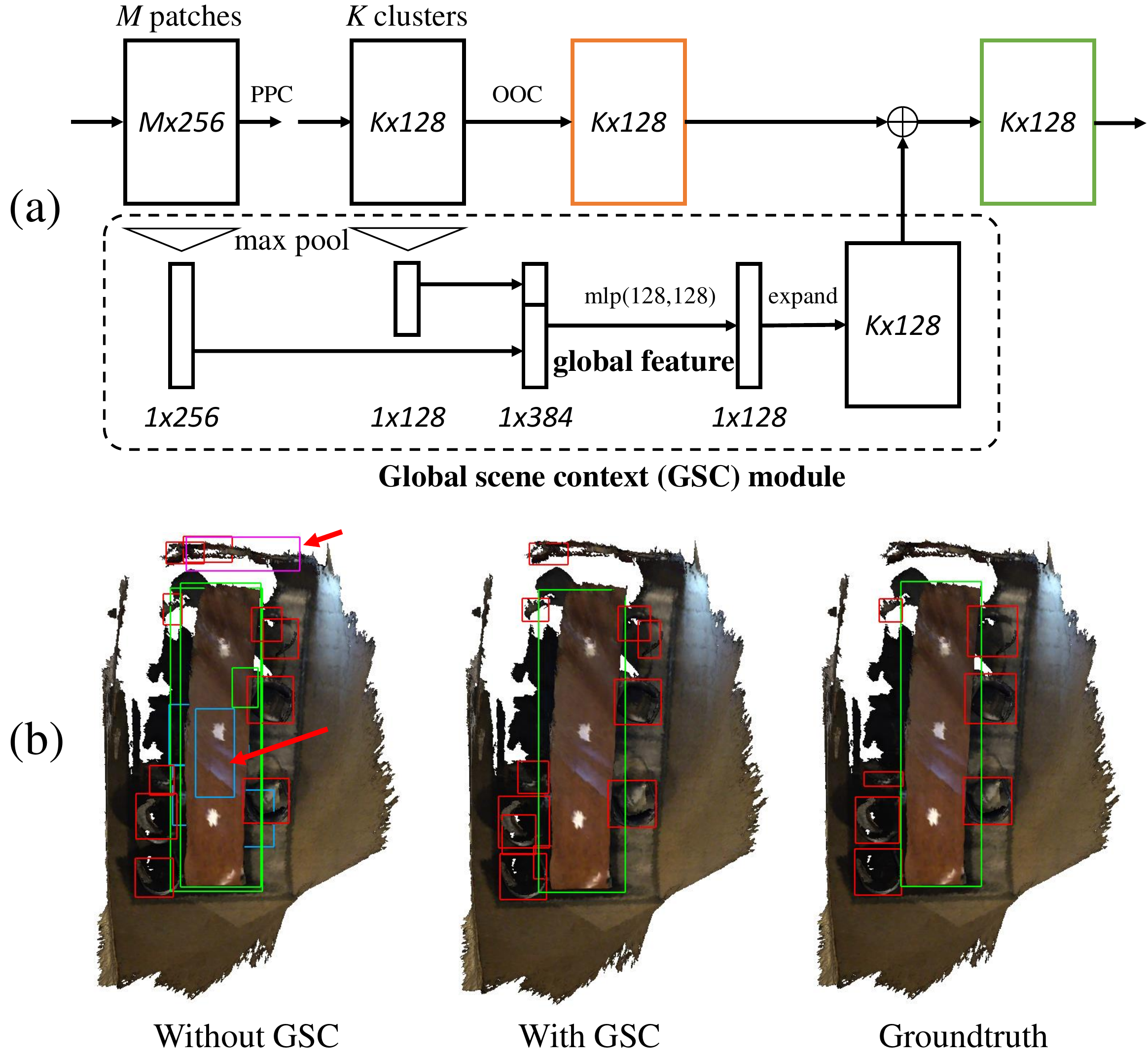}
  \caption{(a) Architecture details of the proposed GSC module with multi-scale feature fusion.
  (b) Comparison of results with and without the GSC module.
  \label{fig:gsc}}
\end{figure}

\subsection{GSC Module}
\label{GSC}
The whole point cloud usually contains rich scene contextual information which can help enhance the object detection accuracy.
For example, it would be highly possible that a chair rather than a toilet is identified when the whole scene is a dining room rather than a bathroom.
Therefore, we regard the information about the whole scene as the third level context, i.e., global scene context (GSC).
Inspired by the idea of scene context extraction in~\cite{liu2018structure}, we propose the GSC module (the green module in Fig.~\ref{fig:network}) to leverage the global scene context information to improve feature representation for 3D bounding box proposal and object classification, without explicit supervision of scenes.

The GSC module is designed to capture the global scene contextual information by introducing a global scene feature extraction branch.
Specifically, we create a new branch with the input from the patch and object levels, concatenating the features at layers before applying self attention in PPC and OOC.
As shown in Fig.~\ref{fig:gsc}(a), at the two layers each row represents a point patch $\mathcal{P} \in \textbf{{P}}=\left\{\mathcal{P}_{1}, \mathcal{P}_{2}, \dots, \mathcal{P}_{M}\right\}$ or an object candidate $\mathcal{C} \in \mathcal{\textbf{C}}=\left\{\mathcal{C}_{1}, \mathcal{C}_{2}, \dots, \mathcal{C}_{K}\right\}$, where
$M$ and $K$ are the numbers of the sampled point patches and clusters, respectively.
Max-pooling is first applied to get two vectors (i.e., the patch vector and the cluster vector), combining information from all the point patches and object candidates.
Following the idea of multi-scale feature fusion in the contextual modeling strategy of 2D detectors, these two vectors are then concatenated to form a global feature vector.
An MLP layer is applied to further aggregate global information, and the output
is subsequently expanded and combined with the output feature map of the OOC module.
This multi-scale feature fusion procedure can be summarized as:
\begin{equation}\label{OOC_formula}
  \mathcal{\textbf{C}}_{new}=MLP([\max (\mathcal{\textbf{C}});\max (\mathcal{\textbf{P}})]) + \mathcal{\textbf{C}}_{OOC}
\end{equation}

In this way, the inference of the final 3D bounding boxes and the object classes will consider the compatibility with the scene context, which makes the final prediction more reliable under the effect of global cues.
As shown in Fig.~\ref{fig:gsc}(b), the GSC module effectively reduces false detection in the scene.

\section{Results and Discussions}
\label{result}

\begin{table}
\centering
\setlength{\tabcolsep}{0.5mm}{
\begin{tabular}{c|c|cc}
\hline
              & Input      & mAP@0.25 & mAP@0.5 \\ \hline
DSS~\cite{song2016deep}           & Geo+RGB    & 15.2     & 6.8     \\
MRCNN 2D-3D~\cite{he2017mask}  &  Geo+RGB     & 17.3  & 10.5    \\
F-PointNet~\cite{qi2018frustum}    & Geo+RGB    & 19.8     & 10.8    \\
GSPN~\cite{yi2019gspn}          & Geo+RGB    & 30.6     & 17.7    \\
3D-SIS~\cite{hou20193d}        & Geo+5views & 40.2     & 22.5    \\ \hline
3D-SIS~\cite{hou20193d}        & Geo only    &   25.4   &  14.6   \\
VoteNet~\cite{qi2019deep}       & Geo only    &   58.6  &   33.5  \\ \hline
MLCVNet(ours) & Geo only    &    \textbf{64.5 }     &    \textbf{41.4}     \\ \hline
\end{tabular}
}
\smallskip
\caption{Performance comparison on ScanNetV2 validation set.
\label{tab:scannet_evaluate}}
\end{table}

\renewcommand\arraystretch{1.2}
\begin{table*}
\centering
\resizebox{\textwidth}{!}
{
\begin{tabular}{c|cccccccccccccccccc|c}
\hline
              & wind           & bed            & cntr           & sofa          & tabl          & showr          & ofurn          & sink           & pic            & chair          & desk           & curt           & fridge         & door           & toil           & bkshf          & bath           & cab            & mAP            \\ \hline
3DSIS5views   & 10.88          & 69.71          & 10.00             & 71.81         & 36.06         & 35.96          & 16.2           & 42.98          & 0.00              & 66.15          & 46.93          & 14.06          & 53.76          & 30.64          & 87.6           & 27.34          & 84.3           & 19.76          & 40.23          \\
3DSISGeo      & 2.79           & 63.14          & 6.92           & 46.33         & 26.91         & 12.17          & 7.05           & 22.87          & 0.00              & 65.98          & 33.34          & 2.47           & 10.42          & 7.95           & 74.51          & 2.3            & 58.66          & 12.75          & 25.36          \\
VoteNet       & 38.1           & 87.92          & 56.13          & \textbf{89.62}         & 58.77         & 57.13          & 37.2           & 54.7           & 7.83           & 88.71          & 71.69          & 47.23          & 45.37          & 47.32          & 94.94          & 44.62          & \textbf{92.11}          & 36.27          & 58.65          \\ \hline
MLCVNet(ours) & \textbf{46.98} & \textbf{88.48} & \textbf{63.94} & 87.4 & \textbf{63.50} & \textbf{65.91} & \textbf{47.89} & \textbf{59.18} & \textbf{11.94} & \textbf{89.98} & \textbf{76.05} & \textbf{56.72} & \textbf{60.86} & \textbf{56.93} & \textbf{98.33} & \textbf{56.94} & 87.22 & \textbf{42.45} & \textbf{64.48} \\ \hline
\end{tabular}
}
\smallskip
\caption{Per-category evaluation on ScanNetV2, evaluated with mAP$@0.25$ IoU.
\label{tab:scannet_detail}}
\end{table*}

\begin{figure*}[!t]
 \centering
 \includegraphics[width=0.9\linewidth]{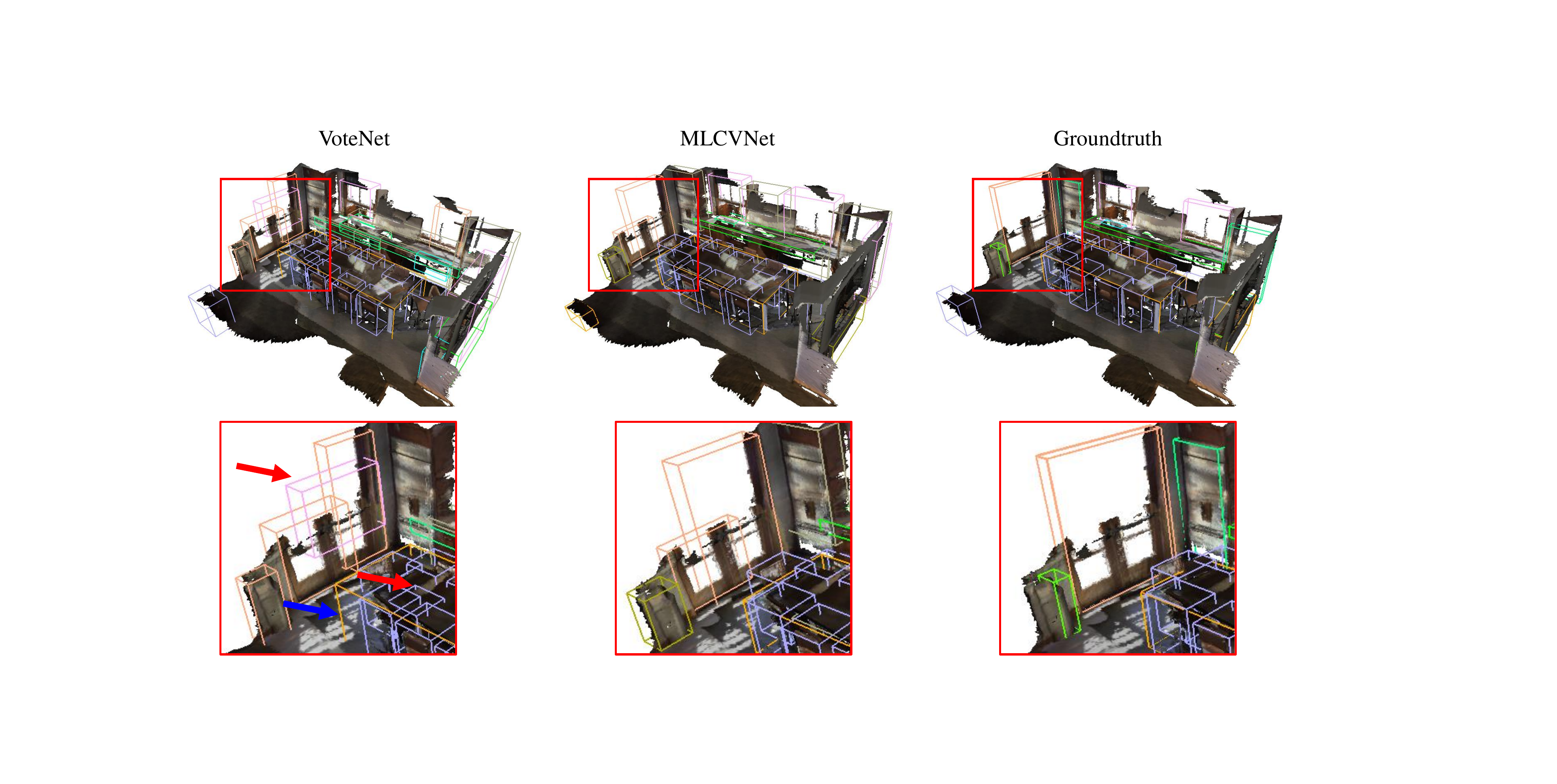}
 \caption{Qualitative comparison results of 3D object detection in ScanNetV2.
 Our multi-level contextual information analysis strategy enables more reasonable and accurate detection.
 \textit{Color is for depiction, not used for detection.}
 \label{fig:scannet_results}}
\end{figure*}

\subsection{Dataset}
\label{subsection:dataset}
We evaluate our approach on SUN RGB-D~\cite{song2015sun} and ScanNet~\cite{dai2017scannet} datasets.
SUN RGB-D is a well-known public RGB-D image dataset of indoor scenes, consisting of 10,335 frames with 3D object bounding box annotations.
Over 64,000 3D bounding boxes are given in the entire dataset.
As described in~\cite{zhang2017deepcontext}, these scenes were mostly taken from household environments with strong context.
The occlusion problem is quite severe in SUN RGB-D dataset.
Sometimes, it is even difficult for humans to recognize the objects in the scene when merely a 3D point cloud is given without any color information.
Thus, it is a challenging dataset for 3D object detection.

ScanNet dataset contains 1513 scanned 3D indoor scenes with densely annotated meshes.
The ground-truth 3D bounding boxes of objects are also provided.
The completeness of scenes in ScanNet makes it an ideal dataset for training our network to learn the contextual information at multiple levels.

\subsection{Training details}
Our network is trained end-to-end using an Adam optimizer and batch size 8.
The base learning rate is set to $0.01$ for ScanNet dataset and $0.001$ for SUN RGB-D dataset.
The network is trained for $220$ epochs on both datasets.
The learning rate decay steps are set to $\{120, 160, 200\}$ for ScanNet, $\{100, 140, 180\}$ for SUN RGB-D, and the decay rates are $\{0.1, 0.1, 0.1\}$.
Training the model until convergence on one RTX 2080 ti GPU takes around 4 hours on ScanNetV2 and 11 hours on SUN RGB-D.
During training we found the mAP result fluctuates within a small range.
Thus, the mAP results reported in the paper are the mean results over three runs.

For parameter size, we check the file sizes of the stored PyTorch models for both our method and VoteNet..
The model size of our network is $13.9MB$, while VoteNet is $11.2MB$.
For training time, VoteNet takes around 40s for 1 epoch with batch size of 8, while ours is around 42s.
For inference time, we infer detection for 1 batch and measure the time.
VoteNet takes around 0.13s, while ours is 0.14s.
The times reported here are all tested on ScanNet dataset.
These show that our method only slightly increases the complexity.

\subsection{Comparisons with the State-of-the-art Methods}
We first evaluate our method on SUN RGB-D dataset using the same 10 most common object categories as in~\cite{qi2019deep}.
Table~\ref{tab:sunrgbd} gives a quantitative comparison of our method with deep sliding shapes (DSS)~\cite{song2016deep}, cloud of gradients (COG)~\cite{ren2016three}, 2D-driven~\cite{lahoud20172d}, F-PointNet~\cite{qi2018frustum} and VoteNet~\cite{qi2019deep}.

\begin{table}
\def\arraystretch{1.1}
\centering
\begin{tabular}{c|ccc|c|c}
\hline
\multirow{3}*{Method} & \multirow{3}*{PPC} & \multirow{3}*{OOC} & \multirow{3}*{GSC} & \multicolumn{2}{c}{mAP$@0.25$}                                    \\
\cline{5-6}
                  ~  &          ~            &            ~          &           ~           & SUN &\multirow{2}*{ScanNet} \\
                  ~  &          ~            &            ~          &           ~           & RGB-D &~ \\
\hline
Baseline                 &                      &                   &   &    57.8  &    59.6     \\ \cline{5-6}
MLCVNet                 & $\surd$              &                      & &    58.6  &    62.2     \\ \cline{5-6}
MLCVNet                 & $\surd$              & $\surd$              & &    59.1  &    63.4     \\ \cline{5-6}
MLCVNet                 & $\surd$              & $\surd$     & $\surd$  &    59.8  &    64.5    \\ \hline
\end{tabular}
\smallskip
\caption{Ablation study on the test dataset. The baseline model is trained by ourselves.\vspace{-5mm}
\label{tab:ablation}}
\end{table}

Remarkably, our method achieves better overall performance
than all the other methods on SUN RGB-D dataset.
The overall mAP (mean average precision) of MLCVNet reaches $59.8\%$ on SUN RGB-D validation set, $2.1\%$ higher than the current state-of-the-art, VoteNet.
The heavy occlusion presented in SUN RGB-D dataset is a challenge for methods (e.g., VoteNet) that consider point patches individually.
However, the utilization of contextual information in MLCVNet helps with the detection of occluded objects with missing parts, which we believe is the reason for the improved detection accuracy.

We also evaluate our MLCVNet against several more competing approaches, MRCNN 2D-3D~\cite{he2017mask}, GSPN~\cite{yi2019gspn} and 3D-SIS~\cite{hou20193d}, on ScanNet benchmark in Table~\ref{tab:scannet_evaluate}.
We report the detection results on both mAP$@0.25$ and mAP$@0.5$.
The mAP$@0.25$ of MLCVNet on ScanNet validation set reaches $64.5\%$ making $5.9$ absolute points improvement over the best competitor VoteNet, and the mAP$@0.50$ is even higher, making $7.9$ points improvement.
The significant improvements confirm the effectiveness of our integration of multi-level contextual information.
Table~\ref{tab:scannet_detail} shows the detailed results at mAP$@0.25$ for each object category in ScanNetV2 dataset.
As can be seen, for some specific categories, such as shower curtain and window, the improvements exceed 8 points.
It is found that plane-like objects, such as door, window, picture and shower curtain, usually get higher improvements.
The reason could be that these objects contain more similar point patches, which can be used by the attention module to complement each other to a great extent.

\begin{figure*}[!t]
 \centering
 \includegraphics[width=0.9\linewidth]{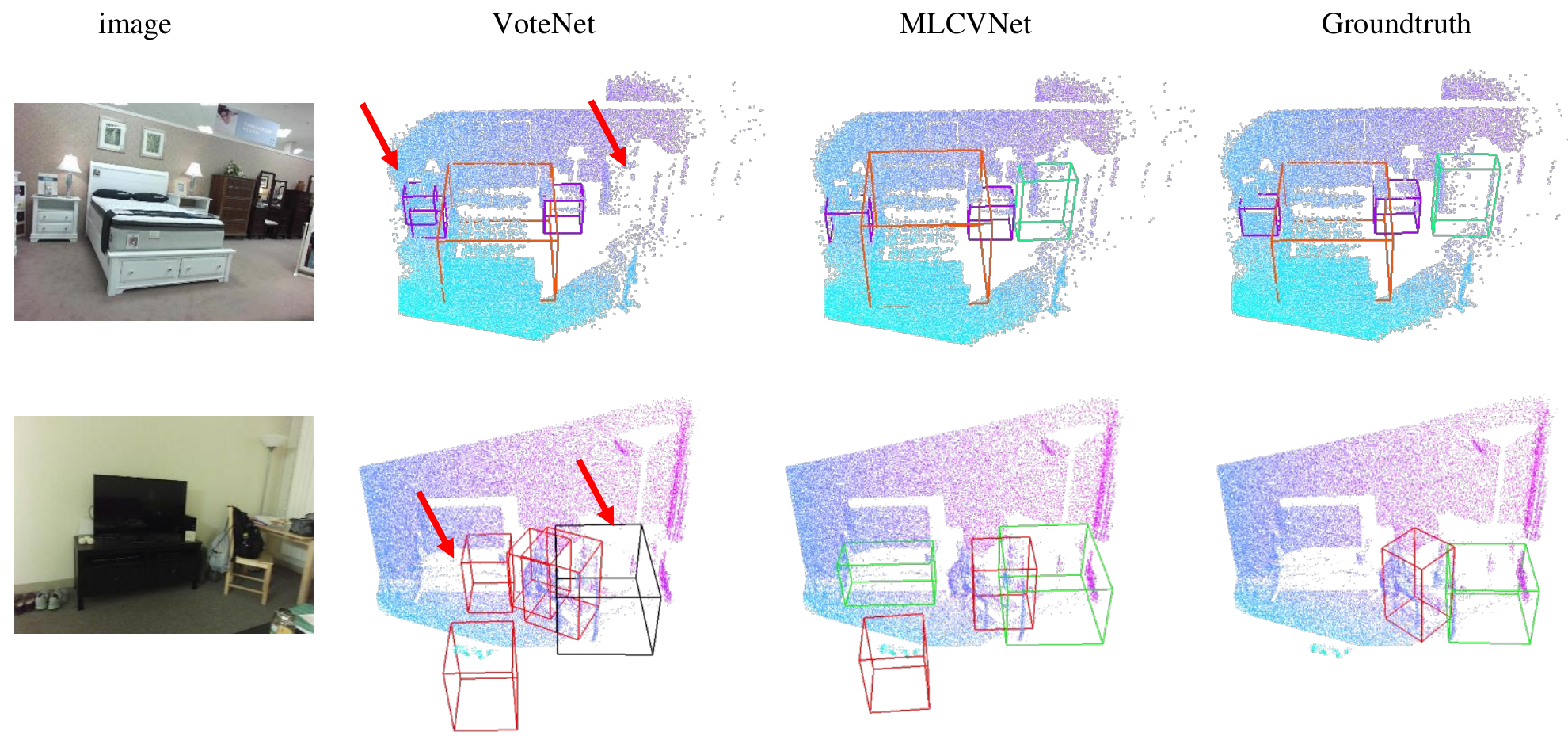}
 \caption{Qualitative results of 3D object detection on SUN RGB-D.
 \label{fig:sun_results}}
\end{figure*}

\subsection{Ablation Study}
To quantitatively evaluate the effectiveness of the proposed contextual sub-modules, we conduct experiments with different combinations of these modules.
The quantitative results are shown in Table~\ref{tab:ablation}.
The baseline method is the VoteNet.
We then add the proposed sub-modules one by one into the baseline model.
Applying the PPC module leads to improvements in mAP$@0.25$ of $0.8$ and $2.6$.
The combination of PPC and OOC modules further improves the evaluation scores to $59.1$ and $63.4$ respectively.
As expected, when equipped with all the three sub-modules, the mAP$@0.25$ of our MLCVNet is boosted up to the highest scores on both datasets.
It can be seen that contextual information captured by the designed sub-modules indeed brings notable improvements over the state-of-the-art method.

\subsection{Qualitative Results}
Fig.~\ref{fig:scannet_results} shows qualitative comparison of the results using MLCVNet and VoteNet for 3D bounding box prediction on ScanNetV2 validation set.
It is observed that the proposed MLCVNet detects more reasonable objects (red arrows), and predicts more precise boxes (blue arrows).
The pink box produced by VoteNet is classified as a window, which is improper to overlap with a door, while our method ensures the compatibility between objects and scenes.
The qualitative comparison results on SUN RGB-D are shown in Fig.~\ref{fig:sun_results}.
As shown, our model is still able to produce high-quality boxes even though the scenes are much occluded and less informative.
As shown in the bedroom example in Fig.~\ref{fig:sun_results}, there are overlaps and missing detection (red arrows) using VoteNet, while our model successfully detects all the objects with good precision compared to the ground-truth.
For the second scene in Fig.~\ref{fig:sun_results}, VoteNet misclassifies the table, produces overlaps, and predicts inaccurate boxes (red arrows), while our model produces much cleaner and more accurate results.
However, it is worth noting that our method may still fail in some predictions, such as the overlapped windows in the red square in Fig.~\ref{fig:scannet_results}.
Therefore, there is still room for improvements on 3D bounding box prediction when dealing with complicated scenes.

\section{Conclusions}
\label{conclusion}
In this paper, we propose a novel network that integrates contextual information at multiple levels into 3D object detection.
We make use of self-attention mechanism and multi-scale feature fusion to model the multi-level contextual information, and propose three sub-modules. The PPC module encodes the relationships between point patches, the OOC module captures the contextual information of object candidates, and the GSC module aggregates the global scene context.
Ablation studies demonstrate the effectiveness of the proposed contextual sub-modules to improve the detection accuracy. Quantitative and qualitative experiments further demonstrate that our architecture successfully improves the performance of 3D object detection.

\textbf{Future work.}
Contextual information analysis in 3D object detection still offers huge space for exploration.
For example, to enhance the global scene context constraint, one possible way is to use the global feature in the GSC module to predict scene types as an auxiliary learning task, which can explicitly supervise the global feature representation.
Another direction would be a more effective mechanism to encode the contextual information as in~\cite{hu2018relation}.


\section*{Acknowledgment}
This work was supported in part by National Natural Science Foundation of China under Grant (61772267, 61572507, 61532003, 61622212), the Fundamental Research Funds for the Central Universities under Grant NE2016004, the National Key Research and Development Program of China (No. 2018AAA0102200) and the Natural Science Foundation of Jiangsu Province under Grant BK20190016.

{\small
\bibliographystyle{unsrt}
\bibliography{egbib}

\begin{thebibliography}{10}

\bibitem{song2016deep}
Shuran Song and Jianxiong Xiao.
\newblock Deep sliding shapes for amodal {3D} object detection in {RGB-D}
  images.
\newblock In {\em Proceedings of the IEEE Conference on Computer Vision and
  Pattern Recognition}, pages 808--816, 2016.

\bibitem{mccormac2018fusion++}
John McCormac, Ronald Clark, Michael Bloesch, Andrew Davison, and Stefan
  Leutenegger.
\newblock Fusion++: Volumetric object-level {SLAM}.
\newblock In {\em 2018 International Conference on 3D Vision (3DV)}, pages
  32--41. IEEE, 2018.

\bibitem{wang2019densefusion}
Chen Wang, Danfei Xu, Yuke Zhu, Roberto Mart{\'\i}n-Mart{\'\i}n, Cewu Lu,
  Li~Fei-Fei, and Silvio Savarese.
\newblock {DenseFusion}: {6D} object pose estimation by iterative dense fusion.
\newblock In {\em Proceedings of the IEEE Conference on Computer Vision and
  Pattern Recognition}, pages 3343--3352, 2019.

\bibitem{qi2017pointnet}
Charles~R Qi, Hao Su, Kaichun Mo, and Leonidas~J Guibas.
\newblock {PointNet}: Deep learning on point sets for {3D} classification and
  segmentation.
\newblock In {\em Proceedings of the IEEE Conference on Computer Vision and
  Pattern Recognition}, pages 652--660, 2017.

\bibitem{qi2017pointnet++}
Charles~Ruizhongtai Qi, Li~Yi, Hao Su, and Leonidas~J Guibas.
\newblock {PointNet++}: Deep hierarchical feature learning on point sets in a
  metric space.
\newblock In {\em Advances in neural information processing systems}, pages
  5099--5108, 2017.

\bibitem{hou20193d}
Ji~Hou, Angela Dai, and Matthias Nie{\ss}ner.
\newblock 3d-sis: {3D} semantic instance segmentation of {RGB-D} scans.
\newblock In {\em Proceedings of the IEEE Conference on Computer Vision and
  Pattern Recognition}, pages 4421--4430, 2019.

\bibitem{qi2019deep}
Charles~R Qi, Or~Litany, Kaiming He, and Leonidas~J Guibas.
\newblock Deep {Hough} voting for {3D} object detection in point clouds.
\newblock {\em arXiv preprint arXiv:1904.09664}, 2019.

\bibitem{hu2018relation}
Han Hu, Jiayuan Gu, Zheng Zhang, Jifeng Dai, and Yichen Wei.
\newblock Relation networks for object detection.
\newblock In {\em Proceedings of the IEEE Conference on Computer Vision and
  Pattern Recognition}, pages 3588--3597, 2018.

\bibitem{yu2016role}
Ruichi Yu, Xi~Chen, Vlad~I Morariu, and Larry~S Davis.
\newblock The role of context selection in object detection.
\newblock {\em arXiv preprint arXiv:1609.02948}, 2016.

\bibitem{zhang2019co}
Hang Zhang, Han Zhang, Chenguang Wang, and Junyuan Xie.
\newblock Co-occurrent features in semantic segmentation.
\newblock In {\em Proceedings of the IEEE Conference on Computer Vision and
  Pattern Recognition}, pages 548--557, 2019.

\bibitem{fu2019dual}
Jun Fu, Jing Liu, Haijie Tian, Yong Li, Yongjun Bao, Zhiwei Fang, and Hanqing
  Lu.
\newblock Dual attention network for scene segmentation.
\newblock In {\em Proceedings of the IEEE Conference on Computer Vision and
  Pattern Recognition}, pages 3146--3154, 2019.

\bibitem{zhang2014panocontext}
Yinda Zhang, Shuran Song, Ping Tan, and Jianxiong Xiao.
\newblock Panocontext: A whole-room {3D} context model for panoramic scene
  understanding.
\newblock In {\em European conference on computer vision}, pages 668--686.
  Springer, 2014.

\bibitem{zhang2017deepcontext}
Yinda Zhang, Mingru Bai, Pushmeet Kohli, Shahram Izadi, and Jianxiong Xiao.
\newblock Deepcontext: Context-encoding neural pathways for {3D} holistic scene
  understanding.
\newblock In {\em Proceedings of the IEEE International Conference on Computer
  Vision}, pages 1192--1201, 2017.

\bibitem{zhang2019pcan}
Wenxiao Zhang and Chunxia Xiao.
\newblock {PCAN}: {3D} attention map learning using contextual information for
  point cloud based retrieval.
\newblock In {\em Proceedings of the IEEE Conference on Computer Vision and
  Pattern Recognition}, pages 12436--12445, 2019.

\bibitem{krizhevsky2012imagenet}
Alex Krizhevsky, Ilya Sutskever, and Geoffrey~E Hinton.
\newblock Imagenet classification with deep convolutional neural networks.
\newblock In {\em Advances in neural information processing systems}, pages
  1097--1105, 2012.

\bibitem{girshick2015fast}
Ross Girshick.
\newblock Fast {R-CNN}.
\newblock In {\em Proceedings of the IEEE international conference on computer
  vision}, pages 1440--1448, 2015.

\bibitem{ren2015faster}
Shaoqing Ren, Kaiming He, Ross Girshick, and Jian Sun.
\newblock Faster {R-CNN}: Towards real-time object detection with region
  proposal networks.
\newblock In {\em Advances in neural information processing systems}, pages
  91--99, 2015.

\bibitem{nan2012search}
Liangliang Nan, Ke~Xie, and Andrei Sharf.
\newblock A search-classify approach for cluttered indoor scene understanding.
\newblock {\em ACM Transactions on Graphics (TOG)}, 31(6):137, 2012.

\bibitem{li2015database}
Yangyan Li, Angela Dai, Leonidas Guibas, and Matthias Nie{\ss}ner.
\newblock Database-assisted object retrieval for real-time {3D} reconstruction.
\newblock In {\em Computer Graphics Forum}, volume~34, pages 435--446. Wiley
  Online Library, 2015.

\bibitem{wang2016cluttered}
Jun Wang, Qian Xie, Yabin Xu, Laishui Zhou, and Nan Ye.
\newblock Cluttered indoor scene modeling via functional part-guided graph
  matching.
\newblock {\em Computer Aided Geometric Design}, 43:82--94, 2016.

\bibitem{wang2017cnn}
Peng-Shuai Wang, Yang Liu, Yu-Xiao Guo, Chun-Yu Sun, and Xin Tong.
\newblock {O-CNN}: Octree-based convolutional neural networks for {3D} shape
  analysis.
\newblock {\em ACM Transactions on Graphics (TOG)}, 36(4):72, 2017.

\bibitem{li2018pointcnn}
Yangyan Li, Rui Bu, Mingchao Sun, Wei Wu, Xinhan Di, and Baoquan Chen.
\newblock {PointCNN}: Convolution on x-transformed points.
\newblock In {\em Advances in Neural Information Processing Systems}, pages
  820--830, 2018.

\bibitem{atzmon2018point}
Matan Atzmon, Haggai Maron, and Yaron Lipman.
\newblock Point convolutional neural networks by extension operators.
\newblock {\em arXiv preprint arXiv:1803.10091}, 2018.

\bibitem{chen2016monocular}
Xiaozhi Chen, Kaustav Kundu, Ziyu Zhang, Huimin Ma, Sanja Fidler, and Raquel
  Urtasun.
\newblock Monocular {3D} object detection for autonomous driving.
\newblock In {\em Proceedings of the IEEE Conference on Computer Vision and
  Pattern Recognition}, pages 2147--2156, 2016.

\bibitem{chen2017multi}
Xiaozhi Chen, Huimin Ma, Ji~Wan, Bo~Li, and Tian Xia.
\newblock Multi-view 3d object detection network for autonomous driving.
\newblock In {\em Proceedings of the IEEE Conference on Computer Vision and
  Pattern Recognition}, pages 1907--1915, 2017.

\bibitem{lahoud20172d}
Jean Lahoud and Bernard Ghanem.
\newblock {2D}-driven {3D} object detection in {RGB-D} images.
\newblock In {\em Proceedings of the IEEE International Conference on Computer
  Vision}, pages 4622--4630, 2017.

\bibitem{qi2018frustum}
Charles~R Qi, Wei Liu, Chenxia Wu, Hao Su, and Leonidas~J Guibas.
\newblock Frustum {PointNets} for {3D} object detection from {RGB-D} data.
\newblock In {\em Proceedings of the IEEE Conference on Computer Vision and
  Pattern Recognition}, pages 918--927, 2018.

\bibitem{zhou2018voxelnet}
Yin Zhou and Oncel Tuzel.
\newblock {VoxelNet}: End-to-end learning for point cloud based {3D} object
  detection.
\newblock In {\em Proceedings of the IEEE Conference on Computer Vision and
  Pattern Recognition}, pages 4490--4499, 2018.

\bibitem{yang2019learning}
Bo~Yang, Jianan Wang, Ronald Clark, Qingyong Hu, Sen Wang, Andrew Markham, and
  Niki Trigoni.
\newblock Learning object bounding boxes for {3D} instance segmentation on
  point clouds.
\newblock {\em arXiv preprint arXiv:1906.01140}, 2019.

\bibitem{shi2019pointrcnn}
Shaoshuai Shi, Xiaogang Wang, and Hongsheng Li.
\newblock {PointRCNN}: {3D} object proposal generation and detection from point
  cloud.
\newblock In {\em Proceedings of the IEEE Conference on Computer Vision and
  Pattern Recognition}, pages 770--779, 2019.

\bibitem{yi2019gspn}
Li~Yi, Wang Zhao, He~Wang, Minhyuk Sung, and Leonidas~J Guibas.
\newblock {GSPN}: Generative shape proposal network for {3D} instance
  segmentation in point cloud.
\newblock In {\em Proceedings of the IEEE Conference on Computer Vision and
  Pattern Recognition}, pages 3947--3956, 2019.

\bibitem{leibe2004combined}
Bastian Leibe, Ales Leonardis, and Bernt Schiele.
\newblock Combined object categorization and segmentation with an implicit
  shape model.
\newblock In {\em Workshop on statistical learning in computer vision, ECCV},
  volume~2, page~7, 2004.

\bibitem{mottaghi2014role}
Roozbeh Mottaghi, Xianjie Chen, Xiaobai Liu, Nam-Gyu Cho, Seong-Whan Lee, Sanja
  Fidler, Raquel Urtasun, and Alan Yuille.
\newblock The role of context for object detection and semantic segmentation in
  the wild.
\newblock In {\em Proceedings of the IEEE Conference on Computer Vision and
  Pattern Recognition}, pages 891--898, 2014.

\bibitem{liu2018structure}
Yong Liu, Ruiping Wang, Shiguang Shan, and Xilin Chen.
\newblock Structure inference net: Object detection using scene-level context
  and instance-level relationships.
\newblock In {\em Proceedings of the IEEE conference on computer vision and
  pattern recognition}, pages 6985--6994, 2018.

\bibitem{deng2018ppfnet}
Haowen Deng, Tolga Birdal, and Slobodan Ilic.
\newblock Ppfnet: Global context aware local features for robust {3D} point
  matching.
\newblock In {\em Proceedings of the IEEE Conference on Computer Vision and
  Pattern Recognition}, pages 195--205, 2018.

\bibitem{engelmann2017exploring}
Francis Engelmann, Theodora Kontogianni, Alexander Hermans, and Bastian Leibe.
\newblock Exploring spatial context for {3D} semantic segmentation of point
  clouds.
\newblock In {\em Proceedings of the IEEE International Conference on Computer
  Vision}, pages 716--724, 2017.

\bibitem{ye20183d}
Xiaoqing Ye, Jiamao Li, Hexiao Huang, Liang Du, and Xiaolin Zhang.
\newblock {3D} recurrent neural networks with context fusion for point cloud
  semantic segmentation.
\newblock In {\em Proceedings of the European Conference on Computer Vision
  (ECCV)}, pages 403--417, 2018.

\bibitem{hu2018semantic}
Shi-Min Hu, Jun-Xiong Cai, and Yu-Kun Lai.
\newblock Semantic labeling and instance segmentation of {3D} point clouds
  using patch context analysis and multiscale processing.
\newblock {\em IEEE transactions on visualization and computer graphics}, 2018.

\bibitem{shi2019hierarchy}
Yifei Shi, Angel~X Chang, Zhelun Wu, Manolis Savva, and Kai Xu.
\newblock Hierarchy denoising recursive autoencoders for {3D} scene layout
  prediction.
\newblock In {\em Proceedings of the IEEE Conference on Computer Vision and
  Pattern Recognition}, pages 1771--1780, 2019.

\bibitem{vaswani2017attention}
Ashish Vaswani, Noam Shazeer, Niki Parmar, Jakob Uszkoreit, Llion Jones,
  Aidan~N Gomez, {\L}ukasz Kaiser, and Illia Polosukhin.
\newblock Attention is all you need.
\newblock In {\em Advances in neural information processing systems}, pages
  5998--6008, 2017.

\bibitem{hu2018squeeze}
Jie Hu, Li~Shen, and Gang Sun.
\newblock Squeeze-and-excitation networks.
\newblock In {\em Proceedings of the IEEE conference on computer vision and
  pattern recognition}, pages 7132--7141, 2018.

\bibitem{xie2018attentional}
Saining Xie, Sainan Liu, Zeyu Chen, and Zhuowen Tu.
\newblock Attentional shapecontextnet for point cloud recognition.
\newblock In {\em Proceedings of the IEEE Conference on Computer Vision and
  Pattern Recognition}, pages 4606--4615, 2018.

\bibitem{paigwar2019attentional}
Anshul Paigwar, Ozgur Erkent, Christian Wolf, and Christian Laugier.
\newblock Attentional {PointNet} for {3D}-object detection in point clouds.
\newblock In {\em Proceedings of the IEEE Conference on Computer Vision and
  Pattern Recognition Workshops}, pages 0--0, 2019.

\bibitem{yue2018compact}
Kaiyu Yue, Ming Sun, Yuchen Yuan, Feng Zhou, Errui Ding, and Fuxin Xu.
\newblock Compact generalized non-local network.
\newblock In {\em Advances in Neural Information Processing Systems}, pages
  6510--6519, 2018.

\bibitem{wang2013learning}
Tao Wang, Xuming He, and Nick Barnes.
\newblock Learning structured {Hough} voting for joint object detection and
  occlusion reasoning.
\newblock In {\em Proceedings of the IEEE Conference on Computer Vision and
  Pattern Recognition}, pages 1790--1797, 2013.

\bibitem{chen2018context}
Zhe Chen, Shaoli Huang, and Dacheng Tao.
\newblock Context refinement for object detection.
\newblock In {\em Proceedings of the European Conference on Computer Vision
  (ECCV)}, pages 71--86, 2018.

\bibitem{ren2016three}
Zhile Ren and Erik~B Sudderth.
\newblock Three-dimensional object detection and layout prediction using clouds
  of oriented gradients.
\newblock In {\em Proceedings of the IEEE Conference on Computer Vision and
  Pattern Recognition}, pages 1525--1533, 2016.

\bibitem{he2017mask}
Kaiming He, Georgia Gkioxari, Piotr Doll{\'a}r, and Ross Girshick.
\newblock Mask {R-CNN}.
\newblock In {\em Proceedings of the IEEE international conference on computer
  vision}, pages 2961--2969, 2017.

\bibitem{song2015sun}
Shuran Song, Samuel~P Lichtenberg, and Jianxiong Xiao.
\newblock {SUN} {RGB-D}: A {RGB-D} scene understanding benchmark suite.
\newblock In {\em Proceedings of the IEEE conference on computer vision and
  pattern recognition}, pages 567--576, 2015.

\bibitem{dai2017scannet}
Angela Dai, Angel~X Chang, Manolis Savva, Maciej Halber, Thomas Funkhouser, and
  Matthias Nie{\ss}ner.
\newblock {ScanNet}: Richly-annotated 3d reconstructions of indoor scenes.
\newblock In {\em Proceedings of the IEEE Conference on Computer Vision and
  Pattern Recognition}, pages 5828--5839, 2017.

\end{thebibliography}
}

\end{document}